\documentclass[11pt,a4paper]{article}
\usepackage[hyperref]{emnlp-ijcnlp-2019}
\usepackage{times}
\usepackage{latexsym}
\usepackage{booktabs}
\usepackage{url}
\usepackage{graphicx}
\usepackage{amsmath}
\usepackage{subcaption}
\aclfinalcopy 

\title{Do Sentence Interactions Matter ? Leveraging Sentence Level Representations for Fake News Classification}

\author{Vaibhav, Raghuram Mandyam Annasamy, Eduard Hovy\\
  Language Technologies Institute \\
  School of Computer Science \\
  Carnegie Mellon University \\
  {\tt mysteryvaibhav@gmail.com, rannasam@cs.cmu.edu, hovy@cmu.edu}}

\date{}

\begin{document}
\maketitle
\begin{abstract}
The rising growth of fake news and misleading information through online media outlets demands an automatic method for detecting such news articles. Of the few limited works which differentiate between trusted vs other types of news article (satire, propaganda, hoax), none of them model sentence interactions within a document. We observe an interesting pattern in the way sentences interact with each other across different kind of news articles. To capture this kind of information for long news articles, we propose a graph neural network-based model which does away with the need of feature engineering for fine grained fake news classification. Through experiments, we show that our proposed method beats strong neural baselines and achieves state-of-the-art accuracy on existing datasets. Moreover, we establish the generalizability of our model by evaluating its performance in out-of-domain scenarios. Code is available at \url{https://github.com/MysteryVaibhav/fake_news_semantics}.
\end{abstract}

\section{Introduction}
In today's day and age of social media, there are ample opportunities for fake news production, dissemination and consumption. \citet{rashkin2017truth} break down fake news into three categories, hoax, propaganda and satire. A hoax article typically tries to convince the reader about a cooked-up story while propaganda ones usually mislead the reader into believing a false political or social agenda. \citet{burfoot2009automatic} defines a satirical article as the one which deliberately exposes real-world individuals, organisations and events to ridicule. 

\begin{figure}[t]
\centering
\begin{subfigure}{.25\textwidth}
  \includegraphics[width=3.5cm, height=2.6cm]{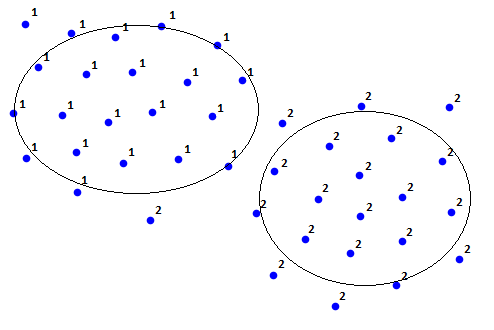}
  \subcaption{Trusted}
\end{subfigure}%
\begin{subfigure}{.2\textwidth}
  \includegraphics[width=3.5cm, height=2.6cm]{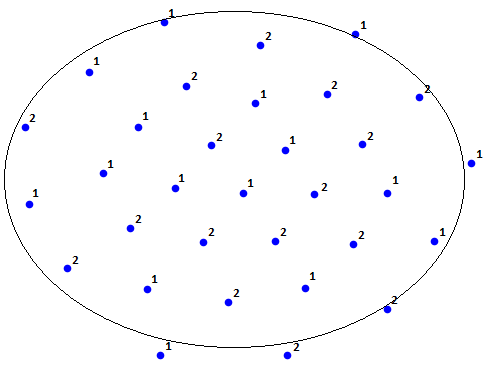}
  \subcaption{Satirical}
\end{subfigure}%
\caption{TSNE visualization~\cite{van2014accelerating} of sentence embeddings obtained using BERT~\cite{devlin2018bert} for two kind of news articles from SLN. A point denotes a sentence and the number indicates which paragraph it belonged to in the article.}
\label{fig:example}
\vspace{-1\baselineskip}
\end{figure}

Previous works \cite{rubin2016fake, rashkin2017truth} rely on various linguistic and hand-crafted semantic features for differentiating between news articles. However, none of them try to model the interaction of sentences within the document. We observed a pattern in the way sentences cluster in different kind of news articles. Specifically, satirical articles had a more coherent story and thus all the sentences in the document \textit{seemed} similar to each other. On the other hand, the trusted news articles were also coherent but the similarity between sentences from different parts of the document was not that strong, as depicted in Figure \ref{fig:example}. We believe that the reason for such kind of behaviour is the presence of factual jumps across sections in a trusted document. 

In this work, we propose a graph neural network-based model to classify news articles while capturing the interaction of sentences across the document. We present a series of experiments on News Corpus with Varying Reliability dataset~\cite{rashkin2017truth} and Satirical Legitimate News dataset~\cite{rubin2016fake}. Our results demonstrate that the proposed model achieves state-of-the-art performance on these datasets and provides interesting insights. Experiments performed in out-of-domain settings establish the generalizability of our proposed method.

\begin{table*}[t]
\centering{}
\scriptsize
\begin{tabular}{c c c c c}
  \toprule
 Dataset & Trusted (\# Docs) & Satire (\# Docs) & Hoax (\# Docs) & Propaganda (\# Docs)\\
  \midrule
	LUN-train & GN except `APW' and `WPB' (9,995) & The Onion (14,047) & American News (6,942) & Activist Report (17,870)\\
	LUN-test & GN only `APW' and `WPB' (750) & The Borowitz Report, Clickhole (750) & DC Gazette (750) & The Natural News (750)\\
	SLN & The Toronto Star, The NY Times (180) & The Onion, The Beaverton (180) & - & -\\
	RPN & WSJ, NBC, etc (75) & The Onion, The Beaverton, etc (75) & - & -\\
  \bottomrule
\end{tabular}
\caption{Statistics about different dataset sources. GN refers to Gigaword News. \label{tab:stats}}
\end{table*}

\begin{figure*}[t]
\begin{center}
\includegraphics[height=2.7cm,width=16cm]{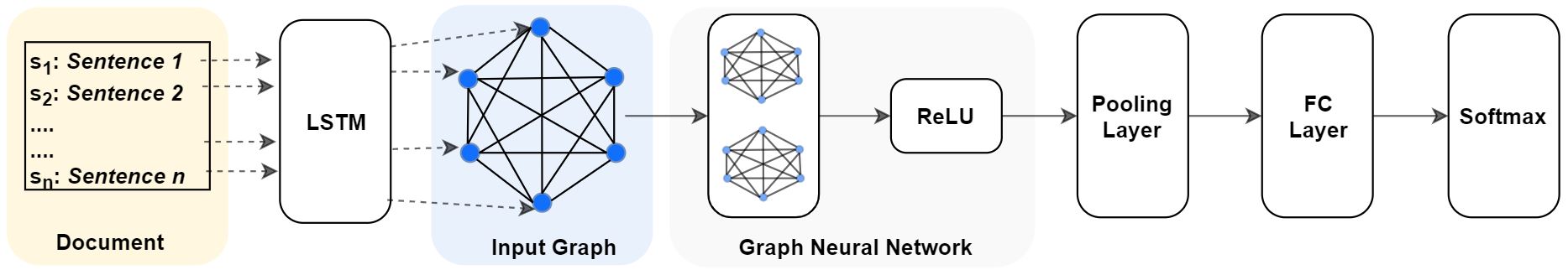}
\end{center}
\caption{Proposed semantic graph neural network based model for fake news classification.}
\label{fig:framework}
\end{figure*}

\section{Related Work}

Satire, according to \citet{simpson2003discourse}, is complicated because it occupies more than one place in the framework for humor, proposed by \citet{ziv1988teaching}: it clearly has an aggressive and social function, and often expresses an intellectual aspect as well. \citet{rubin2016fake} defines news satire as a genre of satire that mimics the format and style of journalistic reporting. Datasets created for the task of identifying satirical news articles from the trusted ones are often constructed by collecting documents from different online sources \cite{rubin2016fake}. \citet{mchardy2019adversarial} hypothesized that this encourages the models to learn characteristics for different publication sources rather than characteristics of satire. In this work, we show that our proposed model generalizes to articles from unseen publication sources.

\citet{rashkin2017truth} extends \citet{rubin2016fake}'s work by offering a quantitative study of linguistic differences found in articles of different types of fake news such as hoax, propaganda and satire. They also proposed predictive models for graded deception across multiple domains. \citet{rashkin2017truth} found that neural methods didn't perform well for this task and proposed to use a Max-Entropy classifier. We show that our proposed neural network based on graph convolutional layers can outperform this model. Recent works by \citet{yang-etal-2017-satirical, de2018attending} show that sophisticated neural models can be used for satirical news detection. To the best of our knowledge, none of the previous works represent individual documents as graphs where the nodes represent the sentences for performing classification using a graph neural network.


\section{Dataset and Baseline}
We use \textbf{SLN}: Satirical and Legitimate News Database~\cite{rubin2016fake}, \textbf{RPN}: Random Political News Dataset~\cite{horne2017just} and \textbf{LUN}: Labeled Unreliable News Dataset~\citet{rashkin2017truth} for our experiments. Table \ref{tab:stats} shows the statistics. Since all of the previous methods on the aforementioned datasets are non-neural, we implement the following neural baselines,
\begin{itemize}
    \item \textbf{CNN:} In this model, we apply a 1-d CNN (Convolutional Neural Network) layer~\cite{kim2014convolutional} with filter size 3 over the word embeddings of the sentences within a document. This is followed by a max-pooling layer to get a single document vector which is passed to a fully connected projection layer to get the logits over output classes.
    \item \textbf{LSTM:} In this model, we encode the document using a LSTM (Long Short-Term Memory) layer~\cite{hochreiter1997long}. We use the hidden state at the last time step as the document vector which is passed to a fully connected projection layer to get the logits over output classes.
    \item \textbf{BERT:} In this model, we extract the sentence vector (representation corresponding to [CLS] token) using BERT (Bidirectional Encoder Representations from Transformers)~\cite{devlin2018bert} for each sentence in the document. We then apply a LSTM layer on the sentence embeddings, followed by a projection layer to make the prediction for each document.
\end{itemize}

\section{Proposed Model}
Capturing sentence interactions in long documents is not feasible using a recurrent network because of the vanishing gradient problem \cite{pascanu2013difficulty}. Thus, we propose a novel way of encoding documents as described in the next subsection. Figure \ref{fig:framework} shows the overall framework of our graph based neural network.

\subsection{Input Representation}
Each document in the corpus is represented as a graph. The nodes of the graph represent the sentences of a document while the edges represent the semantic similarity between a pair of sentences. Representing a document as a fully connected graph allows the model to directly capture the interaction of each sentence with every other sentence in the document. Formally, 
\begin{equation}
    e_{ij} = Similarity(s_i, s_j)
\end{equation}
We initialize the edge scores using BERT \cite{devlin2018bert} finetuned on the semantic textual similarity task\footnote{Task 1 of SemEval-2017} for computing the semantic similarity (SS) between two sentences. Refer to the Supplementary Material for more details regarding the SS model. Note that this representation drops the sentence order information but is better able to capture the interaction between far off sentences within a document.

\subsection{Graph based Neural Networks}
We reformulate the fake news classification problem as a graph classification task, where a graph represents a document. Given a graph $G= (E,S)$ where $E$ is the adjacency matrix and $S$ is the sentence feature matrix. We randomly initialize the word embeddings and use the last hidden state of a LSTM layer as the sentence embedding, shown in Figure \ref{fig:framework}. We experiment with two kinds of graph neural networks,

\subsubsection{Graph Convolution Network (GCN)}
The graph convolutional network \cite{gcn} is a spectral convolutional operation denoted by $f(Z^l, E|W^l)$,
\begin{align}
    Z^{l+1} &= f(Z^l, E|W^l)\\
    f(Z^l, E|W^l) &= \sigma(EZ^lW^l)
\end{align}
Here, $Z^l$ is the output feature corresponding to the nodes after $l^{th}$ convolution. $W^l$ is the parameter associated with the $l^{th}$ layer. We set $Z^0 = S$. Based on the above operation, we can define arbitrarily deep networks. For our experiments, we just use a single layer unless stated otherwise. By default, the adjacency matrix ($E$) is fully connected i.e. all the elements are 1 except the diagonal elements which are all set to 0. We set $E$ based on semantic similarity model in our \textit{GCN + SS} model. For the \textit{GCN + Attn} model, we just add a self attention layer \cite{vaswani2017attention} after the GCN layer and before the pooling layer.

\subsubsection{Graph Attention Network (GAT)}
\citet{gat} introduced graph attention networks to address various shortcomings of GCNs. Most importantly, they enable nodes to attend over their neighborhoods’ features without depending on the graph structure upfront. The key idea is to compute the hidden representations of each node in the graph, by attending over its neighbors, following a self-attention \cite{vaswani2017attention} strategy. 
By default, there is one attention head in the GAT model. For our \textit{GAT + 2 Attn Heads} model, we use two attention heads and concatenate the node embeddings obtained from different heads before passing it to the pooling layer. For a fully connected graph, the GAT model allows every node to attend on every other node and learn the \textit{edge weights}. Thus, initializing the edge weights using the SS model is useless as they are being learned. Mathematical details are provided in the Supplementary Material.

\begin{figure*}[t]
\centering
\begin{subfigure}{.37\textwidth}
\includegraphics[width=7.7cm, height=3.4cm]{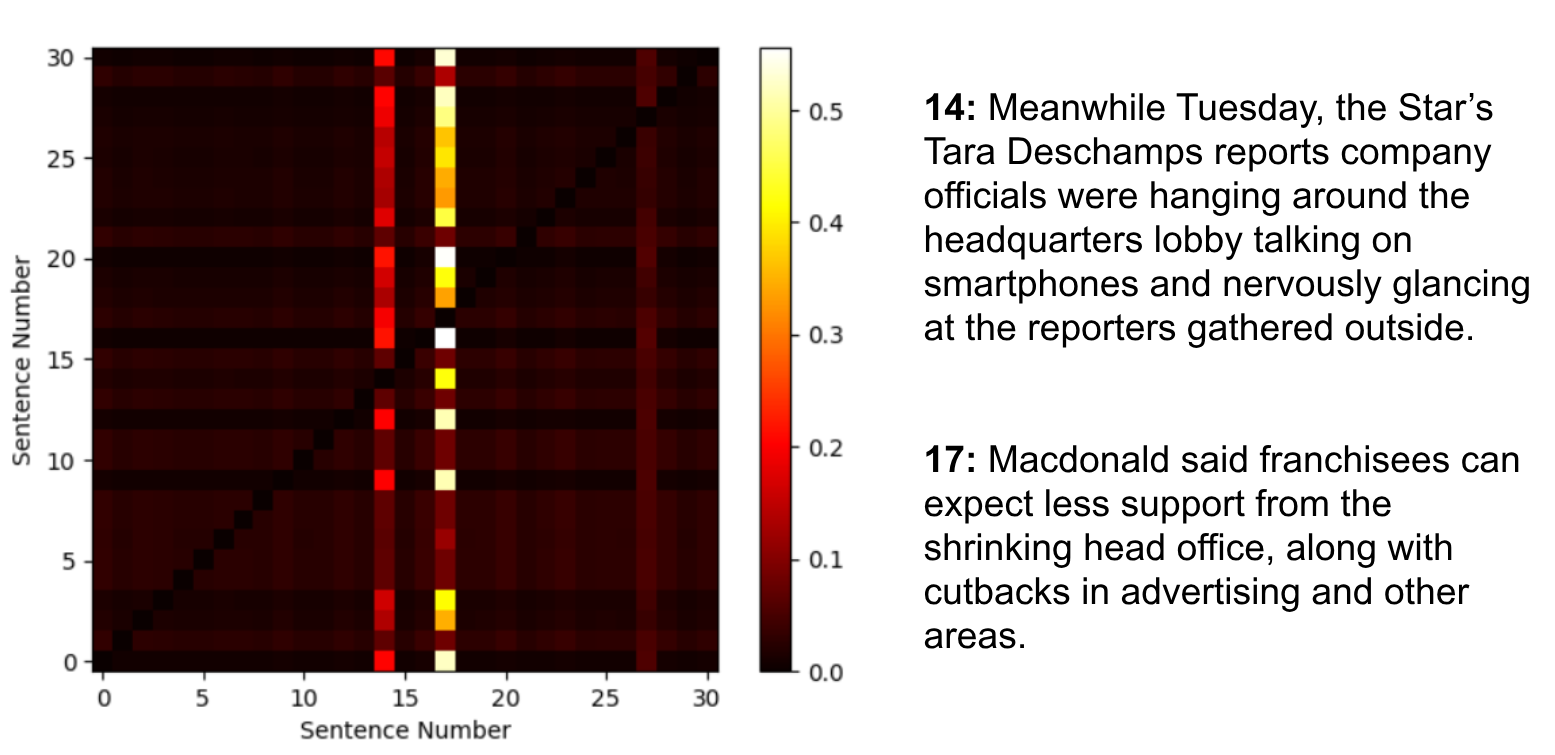}
\end{subfigure}
\hfill
\begin{subfigure}{.50\textwidth}
\includegraphics[width=7.7cm, height=3.4cm]{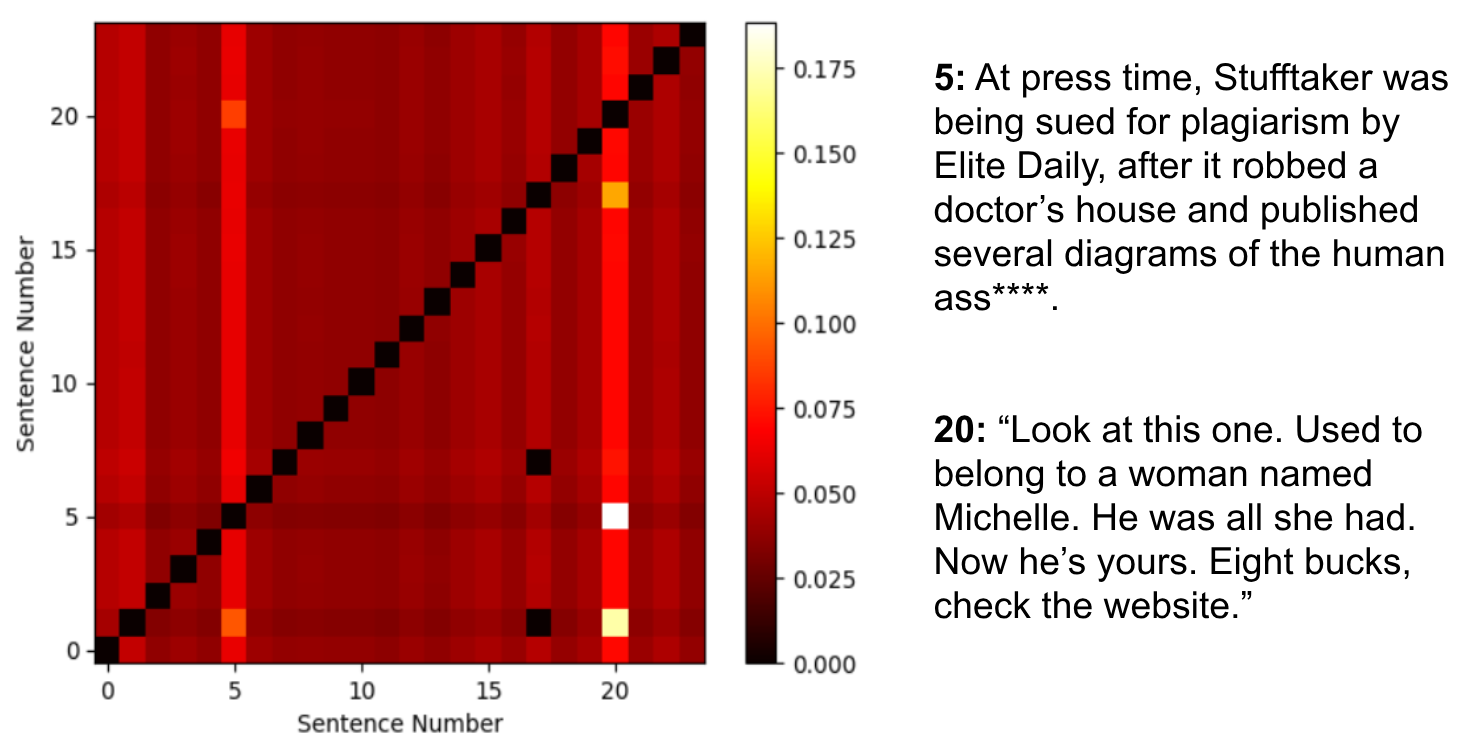}
\end{subfigure}
\caption{Attention heatmaps generated by GAT for 2-way classification. \textbf{Left:} Trusted, \textbf{Right:} Satire.
\label{im:analysis}}
\end{figure*}

\subsection{Hyperparameters}
We use a randomly initialized embedding matrix with 100 dimensions. We use a single layer LSTM to encode the sentences prior to the graph neural networks. All the hidden dimensions used in our networks are set to 100. The node embedding dimension is 32. For GCN and GAT, we set $\sigma$ as \textit{LeakyRelU} with slope 0.2. We train the models for a maximum of 10 epochs and use Adam optimizer with learning rate 0.001. For all the models, we use max-pool for pooling, which is followed by a fully connected projection layer with output nodes equal to the number of classes for classification.

\section{Experimental Setting}
We conduct experiments across various settings and datasets. We report macro-averaged scores in all the settings.\\

\noindent\textbf{2-way classification b/w satire and trusted articles:\label{2-way}} We use the satirical and trusted news articles from LUN-train for training, and from LUN-test as the development set. We evaluate our model on the entire SLN dataset. This is done to emulate a real-world scenario where we want to see the performance of our classifier on an out of domain dataset. We don't use SLN for training purposes because it just contains 360 examples which is too little for training our model and we want to have an unseen test set. The best performing model on SLN is used to evaluate the performance on RPN.\\

\noindent \textbf{4-way classification b/w satire, propaganda, hoax and trusted articles:} We split the LUN-train into a 80:20 split to create our training and development set. We use the LUN-test as our out of domain test set.

\section{Results}

\begin{table}[t]
\centering{}
\small
\begin{tabular}{l c c}
  \toprule
 Model & \bf Precision &\bf Recall\\
  \midrule
CNN & 67.5&	67.5\\
LSTM  &82.2	&81.4\\
BERT  & 78.1&	78.1\\
SoTA* \cite{rubin2016fake} & 88.0 & 82.0\\
\midrule
\multicolumn{3}{c}{Our Models}\\
\midrule
GCN & 85.9	&85.0\\
GCN + SS & 86.4&	86.3\\
GCN + Attn & 87.1	&86.9\\
GCN + Attn + SS & 87.8	&87.8\\
GAT &  86.2	&86.1\\
GAT + 2 Attn Heads& \textbf{89.1}&	\textbf{88.9}\\
  \bottomrule
\end{tabular}
\caption{2-way classification results on SLN. *n-fold cross validation (precision, recall) as reported in SoTA.}
\label{tbl:results:2way}
\vspace{-1\baselineskip}
\end{table}

\begin{table}[t]
\centering{}
\small
\begin{tabular}{l c c}
  \toprule
  \bf Model & \bf LUN-dev & \bf LUN-test\\
  \midrule
CNN & 96.48 & 54.04\\
LSTM & 88.75 & 55.05\\
BERT & 95.07 & 54.87\\
SoTA* \cite{rashkin2017truth} & 91.0 & 65.0\\
\midrule
\multicolumn{3}{c}{Our Models}\\
\midrule
GCN & 96.76 & 65.0\\
GCN + Attn & 97.57 & \textbf{67.08}\\
GAT & 97.28 & 65.51\\
GAT + 2 Attn Heads & \textbf{97.82} & 66.95\\
  \bottomrule
\end{tabular}
\caption{4-way classification results for different models. We only report F1-score following the SoTA paper.}
\label{tbl:results:4way}
\vspace{-1\baselineskip}
\end{table}

Table \ref{tbl:results:2way} shows the quantitative results for the two way classification between satirical and trusted news articles. Our proposed GAT method with 2 attention heads outperforms SoTA. The semantic similarity model does not seem to have much impact on the GCN model, and considering the computing cost, we don't experiment with it for the 4-way classification scenario. Given that we use SLN as an out of domain test set (just one overlapping source, no overlap in articles), whereas the SoTA paper \cite{rubin2016fake} reports a 10-fold cross validation number on SLN. We believe that our results are quite strong, the \textit{GAT + 2 Attn Heads} model achieves an accuracy of 87\% on the entire RPN dataset when used as an out-of-domain test set. The SoTA paper \cite{horne2017just} on RPN reports a 5-fold cross validation accuracy of 91\%. These results indicate the generalizability of our proposed model across datasets.
We also present results of four way classification in Table \ref{tbl:results:4way}. All of our proposed methods outperform SoTA on both the in-domain and out of domain test set.

To further understand the working of our proposed model, we closely inspect the attention maps generated by the GAT model for satirical and trusted news articles for the SLN dataset. From Figure \ref{im:analysis}, we can see that the attention map generated for the trusted news article only focuses on two specific sentence whereas the attention weights are much more distributed in case of a satirical article. Interestingly enough the highlighted sentences in case of the trusted news article were the starting sentence of two different paragraphs in the article indicating the presence of similar sentence clusters within a document. This opens a new avenue for understanding the differences between different kind of text articles for future research.

\section{Conclusion}
This paper introduces a novel way of encoding articles for fake news classification. The intuition behind representing documents as a graph is motivated by the fact that sentences interact differently with each other across different kinds of article. Recurrent networks are unable to maintain long term dependencies in large documents, whereas a fully connected graph captures the interaction between sentences at unit distance. The quantitative result shows the effectiveness of our proposed model and the qualitative results validate our hypothesis about difference in sentence interaction across different articles. Further, we show that our proposed model generalizes to unseen datasets. 

\section*{Acknowledgement}
We would like to thank the AWS Educate program for donating computational GPU resources used in this work. We also appreciate the anonymous reviewers for their insightful comments and suggestions to improve the paper.

\section*{Supplementary Material}
The supplementary material is available\footnote{\url{https://github.com/MysteryVaibhav/fake_news_semantics/blob/master/EMNLP2019_TextGraphs_Supplementary.pdf}} along with the code which provides mathematical details of the GAT model and few additional qualitative results.

\bibliography{emnlp-ijcnlp-2019}

\begin{thebibliography}{17}
\expandafter\ifx\csname natexlab\endcsname\relax\def\natexlab#1{#1}\fi

\bibitem[{Burfoot and Baldwin(2009)}]{burfoot2009automatic}
Clint Burfoot and Timothy Baldwin. 2009.
\newblock Automatic satire detection: Are you having a laugh?
\newblock In \emph{Proceedings of the ACL-IJCNLP 2009 conference short papers},
  pages 161--164. Association for Computational Linguistics.

\bibitem[{De~Sarkar et~al.(2018)De~Sarkar, Yang, and
  Mukherjee}]{de2018attending}
Sohan De~Sarkar, Fan Yang, and Arjun Mukherjee. 2018.
\newblock Attending sentences to detect satirical fake news.
\newblock In \emph{Proceedings of the 27th International Conference on
  Computational Linguistics}, pages 3371--3380.

\bibitem[{Devlin et~al.(2019)Devlin, Chang, Lee, and
  Toutanova}]{devlin2018bert}
Jacob Devlin, Ming-Wei Chang, Kenton Lee, and Kristina Toutanova. 2019.
\newblock \href {https://doi.org/10.18653/v1/N19-1423} {{BERT}: Pre-training of
  deep bidirectional transformers for language understanding}.
\newblock In \emph{Proceedings of the 2019 Conference of the North {A}merican
  Chapter of the Association for Computational Linguistics: Human Language
  Technologies, Volume 1 (Long and Short Papers)}, pages 4171--4186,
  Minneapolis, Minnesota. Association for Computational Linguistics.

\bibitem[{Hochreiter and Schmidhuber(1997)}]{hochreiter1997long}
Sepp Hochreiter and J{\"u}rgen Schmidhuber. 1997.
\newblock Long short-term memory.
\newblock \emph{Neural computation}, 9(8):1735--1780.

\bibitem[{Horne and Adali(2017)}]{horne2017just}
Benjamin~D Horne and Sibel Adali. 2017.
\newblock This just in: fake news packs a lot in title, uses simpler,
  repetitive content in text body, more similar to satire than real news.
\newblock In \emph{Eleventh International AAAI Conference on Web and Social
  Media}.

\bibitem[{Kim(2014)}]{kim2014convolutional}
Yoon Kim. 2014.
\newblock Convolutional neural networks for sentence classification.
\newblock In \emph{Proceedings of the 2014 Conference on Empirical Methods in
  Natural Language Processing (EMNLP)}, pages 1746--1751.

\bibitem[{Kipf and Welling(2017)}]{gcn}
Thomas~N. Kipf and Max Welling. 2017.
\newblock Semi-supervised classification with graph convolutional networks.
\newblock In \emph{International Conference on Learning Representations
  (ICLR)}.

\bibitem[{McHardy et~al.(2019)McHardy, Adel, and
  Klinger}]{mchardy2019adversarial}
Robert McHardy, Heike Adel, and Roman Klinger. 2019.
\newblock \href {https://doi.org/10.18653/v1/N19-1069} {Adversarial training
  for satire detection: Controlling for confounding variables}.
\newblock In \emph{Proceedings of the 2019 Conference of the North {A}merican
  Chapter of the Association for Computational Linguistics: Human Language
  Technologies, Volume 1 (Long and Short Papers)}, pages 660--665, Minneapolis,
  Minnesota. Association for Computational Linguistics.

\bibitem[{Pascanu et~al.(2013)Pascanu, Mikolov, and
  Bengio}]{pascanu2013difficulty}
Razvan Pascanu, Tomas Mikolov, and Yoshua Bengio. 2013.
\newblock On the difficulty of training recurrent neural networks.
\newblock In \emph{International conference on machine learning}, pages
  1310--1318.

\bibitem[{Rashkin et~al.(2017)Rashkin, Choi, Jang, Volkova, and
  Choi}]{rashkin2017truth}
Hannah Rashkin, Eunsol Choi, Jin~Yea Jang, Svitlana Volkova, and Yejin Choi.
  2017.
\newblock Truth of varying shades: Analyzing language in fake news and
  political fact-checking.
\newblock In \emph{Proceedings of the 2017 Conference on Empirical Methods in
  Natural Language Processing}, pages 2931--2937.

\bibitem[{Rubin et~al.(2016)Rubin, Conroy, Chen, and Cornwell}]{rubin2016fake}
Victoria Rubin, Niall Conroy, Yimin Chen, and Sarah Cornwell. 2016.
\newblock Fake news or truth? using satirical cues to detect potentially
  misleading news.
\newblock In \emph{Proceedings of the Second Workshop on Computational
  Approaches to Deception Detection}, pages 7--17.

\bibitem[{Simpson(2003)}]{simpson2003discourse}
Paul Simpson. 2003.
\newblock \emph{On the discourse of satire: Towards a stylistic model of
  satirical humour}, volume~2.
\newblock John Benjamins Publishing.

\bibitem[{Van Der~Maaten(2014)}]{van2014accelerating}
Laurens Van Der~Maaten. 2014.
\newblock Accelerating t-sne using tree-based algorithms.
\newblock \emph{The Journal of Machine Learning Research}, 15(1):3221--3245.

\bibitem[{Vaswani et~al.(2017)Vaswani, Shazeer, Parmar, Uszkoreit, Jones,
  Gomez, Kaiser, and Polosukhin}]{vaswani2017attention}
Ashish Vaswani, Noam Shazeer, Niki Parmar, Jakob Uszkoreit, Llion Jones,
  Aidan~N Gomez, {\L}ukasz Kaiser, and Illia Polosukhin. 2017.
\newblock Attention is all you need.
\newblock In \emph{Advances in neural information processing systems}, pages
  5998--6008.

\bibitem[{Veličković et~al.(2018)Veličković, Cucurull, Casanova, Romero,
  Liò, and Bengio}]{gat}
Petar Veličković, Guillem Cucurull, Arantxa Casanova, Adriana Romero, Pietro
  Liò, and Yoshua Bengio. 2018.
\newblock \href {https://openreview.net/forum?id=rJXMpikCZ} {Graph attention
  networks}.
\newblock In \emph{International Conference on Learning Representations}.

\bibitem[{Yang et~al.(2017)Yang, Mukherjee, and
  Dragut}]{yang-etal-2017-satirical}
Fan Yang, Arjun Mukherjee, and Eduard Dragut. 2017.
\newblock \href {https://doi.org/10.18653/v1/D17-1211} {Satirical news
  detection and analysis using attention mechanism and linguistic features}.
\newblock In \emph{Proceedings of the 2017 Conference on Empirical Methods in
  Natural Language Processing}, pages 1979--1989, Copenhagen, Denmark.
  Association for Computational Linguistics.

\bibitem[{Ziv(1988)}]{ziv1988teaching}
Avner Ziv. 1988.
\newblock Teaching and learning with humor: Experiment and replication.
\newblock \emph{The Journal of Experimental Education}, 57(1):4--15.

\end{thebibliography}
\bibliographystyle{acl_natbib}


\end{document}